\documentclass[11pt,a4paper]{article}
\PassOptionsToPackage{breaklinks}{hyperref}
\usepackage[hyperref]{emnlp-ijcnlp-2019}
\usepackage[utf8]{inputenc}
\usepackage{times}
\usepackage{latexsym}
\usepackage{color, colortbl}
\usepackage{pifont}

\usepackage{graphicx}
\usepackage{multirow}
\usepackage{tabularx}
\usepackage{breakurl}

\usepackage{xr}

\newenvironment{citemize}{\begin{list}{$\bullet$}{\topsep=.1\smallskipamount\itemsep=0pt\parsep=1pt\labelwidth=.5em}}{\end{list}}

\aclfinalcopy 

\title{Evaluating Contextualized Embeddings on 54 Languages\\
in POS Tagging, Lemmatization and Dependency Parsing}

\author{Milan Straka \and Jana Strakov\'{a} \and Jan Haji\v{c}\\
  Charles University \\
  Faculty of Mathematics and Physics \\
  Institute of Formal and Applied Linguistics \\
  {\tt \{strakova,straka,hajic\}@ufal.mff.cuni.cz} \\}

\date{}

\begin{document}
\maketitle

\begin{abstract}
  We present an extensive evaluation of three recently proposed methods for
  contextualized embeddings on $89$ corpora in $54$ languages of the Universal
  Dependencies 2.3 in three tasks: POS tagging, lemmatization, and dependency
  parsing. Employing the BERT, Flair and ELMo as pretrained embedding inputs in
  a strong baseline of UDPipe 2.0, one of the best-performing systems of the
  CoNLL 2018 Shared Task and an overall winner of the EPE 2018, we present
  a one-to-one comparison of the three contextualized word embedding methods,
  as well as a comparison with word2vec-like pretrained embeddings and with
  end-to-end character-level word embeddings. We report state-of-the-art
  results in all three tasks as compared to results on UD 2.2 in the CoNLL 2018
  Shared Task.
\end{abstract}

\section{Introduction}
\label{section:introduction}

We publish a comparison and evaluation of three recently proposed
contextualized word embedding methods: BERT \cite{BERT}, Flair \cite{Akbik} and
ELMo \cite{Peters2018}, in $89$ corpora which have a training set in $54$
languages of the Universal Dependencies 2.3 in three tasks: POS tagging,
lemmatization and dependency parsing.
Our contributions are the following:
\begin{citemize}
  \item Meaningful massive comparative evaluation of BERT
    \cite{BERT}, Flair \cite{Akbik} and ELMo \cite{Peters2018} contextualized
    word embeddings, by adding them as input features to a strong baseline of
    UDPipe 2.0, one of the best performing systems in the CoNLL 2018 Shared
    Task \cite{CoNLL2018} and an overall winner of the EPE 2018 Shared Task
    \cite{EPE2018}.
  \item State-of-the-art results in POS tagging, lemmatization and dependency
    parsing in UD 2.2, the dataset used in CoNLL 2018 Shared Task
    \cite{CoNLL2018}.
  \item We report our best results on UD 2.3. The addition of contextualized
    embeddings improvements range from $25\%$ relative error reduction for
    English treebanks, through $20\%$ relative error reduction for high
    resource languages, to $10\%$ relative error reduction for all UD 2.3
    languages which have a training set.
\end{citemize}

\begin{figure*}[t]
  \begin{center}
    \includegraphics[width=\hsize]{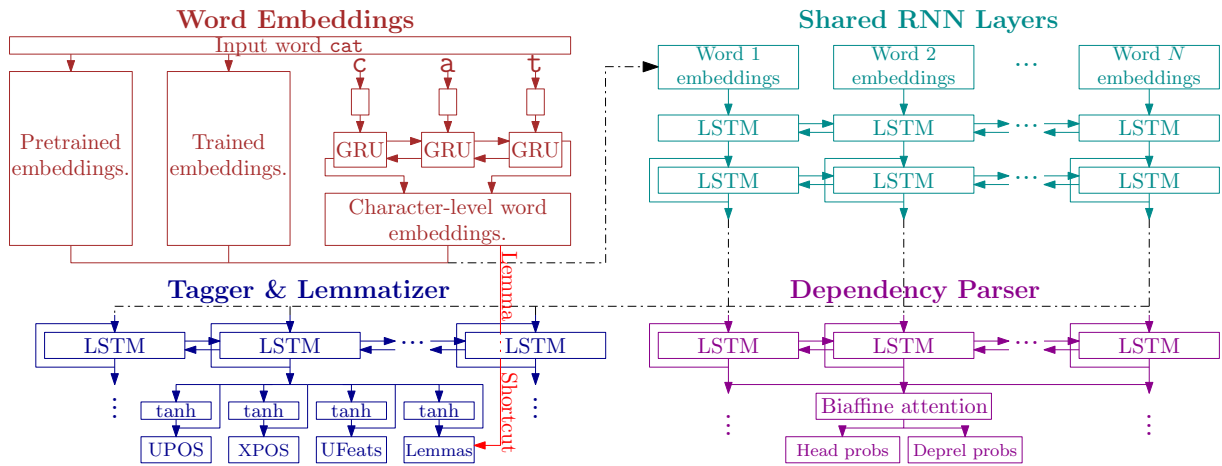}
  \end{center}
  \caption{UDPipe 2.0 architecture overview.}
  \label{figure:udpipe}
\end{figure*}

\section{Related Work}
\label{section:related_work}

A new type of deep contextualized word representation was introduced by
\citet{Peters2018}. The proposed embeddings, called ELMo, were obtained from
internal states of deep bidirectional language model, pretrained on a large
text corpus. \citet{Akbik} introduced analogous contextual string
embeddings called Flair, which were obtained from internal states of
a \emph{character-level} bidirectional language model. The idea of ELMos was
extended by \citet{BERT}, who instead of a bidirectional recurrent language
model employ a Transformer \citep{vaswani:2017} architecture.

The \emph{Universal
Dependencies}\footnote{\scriptsize\url{https://universaldependencies.org/}}
project \citep{ud} seeks to develop cross-linguistically consistent treebank
annotation of morphology and syntax for many languages.  The latest version
UD~2.3~\citep{ud23} consists of $129$ treebanks in $76$ languages,
with $89$ of the treebanks containing a train a set and being freely available.
The annotation consists of UPOS (universal POS tags), XPOS (language-specific
POS tags), Feats (universal morphological features), Lemmas, dependency heads
and universal dependency labels.

In 2017 and 2018, CoNLL Shared Tasks \emph{Multilingual Parsing from Raw Text
to Universal Dependencies} \cite{CoNLL2017,CoNLL2018} were held in order to
stimulate research in multi-lingual POS tagging, lemmatization and dependency
parsing.

The system of \citet{udst18:harbin} is one of the three winners of the CoNLL
2018 Shared Task. The authors employed manually trained ELMo-like contextual
word embeddings, reporting $7.9\%$ error reduction in LAS parsing performance.

\begin{table*}[t]
  \begin{center}
    \begin{tabular}{c|c|c||r|r|r|r|r|r|r|r}
\multicolumn{1}{c|}{WE} & \multicolumn{1}{c|}{CLE} & \multicolumn{1}{c||}{Bert} & \multicolumn{1}{c|}{\kern-.1em UPOS\kern-.1em} & \multicolumn{1}{c|}{XPOS} & \multicolumn{1}{c|}{\kern-.2em UFeats\kern-.2em} & \multicolumn{1}{c|}{\kern-.25em Lemma\kern-.25em} & \multicolumn{1}{c|}{UAS} & \multicolumn{1}{c|}{LAS} & \multicolumn{1}{c|}{\kern-.2em MLAS\kern-.2em} & \multicolumn{1}{c}{\kern-.2em BLEX\kern-.2em}\\\hline\hline
 &  &  & 90.14 & 88.51 & 86.50 & 88.64 & 79.43 & 73.55 & 56.52 & 60.84\\\hline
WE &  &  & 94.91 & 93.51 & 91.89 & 92.10 & 85.98 & 81.73 & 68.47 & 70.64\\\hline
 & CLE &  & 95.75 & 94.69 & 93.43 & 96.24 & 86.99 & 82.96 & 71.06 & 75.78\\\hline
WE & CLE &  & 96.39 & 95.53 & 94.28 & 96.51 & 87.79 & 84.09 & 73.30 & 77.36\\\hline
 &  & Base & 96.35 & 95.08 & 93.56 & 93.29 & 89.31 & 85.69 & 74.11 & 75.45\\\hline
WE &  & Base & 96.62 & 95.54 & 94.08 & 93.77 & 89.49 & 85.96 & 74.94 & 76.27\\\hline
 & CLE & Base & 96.86 & 95.96 & 94.85 & 96.64 & 89.76 & 86.29 & 76.20 & 79.87\\\hline
WE & CLE & Base & \bf 97.00 & \bf 96.17 & \bf 94.97 & \bf 96.66 & \bf 89.81 & \bf 86.42 & \bf 76.54 & \bf 80.04\\\hline
\end{tabular}

  \end{center}
  \caption{BERT Base compared to word embeddings (WE) and character-level word
  embeddings (CLE). Results for $72$ UD 2.3 treebanks with train and development sets
  and non-empty Wikipedia.}
  \label{table:bert}
\end{table*}

\begin{table*}[t]
  \begin{center}
    \begin{tabular}{l|c||r|r|r|r|r|r|r|r}
\multicolumn{1}{c|}{Language} & \multicolumn{1}{c||}{Bert} & \multicolumn{1}{c|}{\kern-.1em UPOS\kern-.1em} & \multicolumn{1}{c|}{XPOS} & \multicolumn{1}{c|}{\kern-.2em UFeats\kern-.2em} & \multicolumn{1}{c|}{\kern-.25em Lemma\kern-.25em} & \multicolumn{1}{c|}{UAS} & \multicolumn{1}{c|}{LAS} & \multicolumn{1}{c|}{\kern-.2em MLAS\kern-.2em} & \multicolumn{1}{c}{\kern-.2em BLEX\kern-.2em}\\\hline\hline
English & Base & \bf 97.38 & \bf 96.97 & 97.22 & \bf 97.71 & \bf 91.09 & \bf 88.22 & \bf 80.48 & \bf 82.38\\\hline
English & Multi & 97.36 & \bf 96.97 & \bf 97.29 & 97.63 & 90.94 & 88.12 & 80.43 & 82.22\\\hline
\hline
Chinese & Base & \bf 97.07 & \bf 96.89 & \bf 99.58 & 99.98 & \bf 90.13 & \bf 86.74 & \bf 79.67 & \bf 83.85\\\hline
Chinese & Multi & 96.27 & 96.25 & 99.37 & \bf 99.99 & 87.58 & 83.96 & 76.26 & 81.04\\\hline
\hline
Japanese & Base & \bf 98.24 & \bf 97.89 & 99.98 & \bf 99.53 & \bf 95.55 & \bf 94.27 & \bf 87.64 & \bf 89.24\\\hline
Japanese & Multi & 98.17 & 97.71 & \bf 99.99 & 99.51 & 95.30 & 93.99 & 87.17 & 88.77\\\hline
\end{tabular}

  \end{center}
  \caption{Comparison of multilingual and language-specific BERT models on
  4 English treebanks (each experiment repeated 3 times), and on Chinese-GSD and Japanese-GSD treebanks.}
  \label{table:bert_langs}
\end{table*}

\section{Methods}
\label{section:methods}

Our \textbf{baseline} is the \textit{UDPipe 2.0} \cite{UDPipe2.0} participant
system from the CoNLL 2018 Shared Task \cite{CoNLL2018}. The system is
available at {\small\url{http://github.com/CoNLL-UD-2018/UDPipe-Future}}.

A graphical overview of the UDPipe 2.0 is shown in Figure~\ref{figure:udpipe}.
In short, UDPipe 2.0 is a multi-task model predicting POS tags,
lemmas and dependency trees jointly. After embedding input words, two shared
bidirectional LSTM~\citep{Hochreiter:1997:LSTM} layers are performed. Then,
tagger and lemmatizer specific bidirectional LSTM layer is executed, with
softmax classifiers processing its output and generating UPOS, XPOS, Feats and
Lemmas. The lemmas are generated by classifying into a set of edit scripts
which process input word form and produce lemmas by performing character-level
edits on the word prefix and suffix. The lemma classifier additionally takes
the character-level word embeddings as input.

Finally, the output of the two shared LSTM layers is processed by
a parser specific bidirectional LSTM layer, whose output is then passed to
a biaffine attention layer \cite{dozat:2016} producing labeled dependency
trees. We refer the readers for detailed treatment of the architecture and
the training procedure to \citet{UDPipe2.0}.

The simplest baseline system uses only end-to-end word embeddings trained
specifically for the task.
Additionally, the UDPipe 2.0 system also employs the following two embeddings:
\begin{citemize}
  \item \textbf{word embeddings (WE):} We use FastText word embeddings
    \cite{FastText} of dimension $300$, which we pretrain for each language
    on Wikipedia using segmentation and tokenization trained from the UD
    data.\footnote{We use {\scriptsize\ttfamily-minCount 5 -epoch 10 -neg 10}
    options and keep at most one million most frequent words.}
  \item \textbf{character-level word embeddings (CLE):} We employ bidirectional
    GRUs of dimension $256$ in line with
    \citet{Ling2015}: we represent every Unicode character with a vector of
    dimension $256$, and concatenate GRU output for forward and reversed word
    characters. The character-level word embeddings are trained together with
    UDPipe network.
\end{citemize}

Optionally, we add pretrained contextual word embeddings as another input
to the neural network. Contrary to finetuning approach used by the BERT
authors \citep{BERT}, we never finetune the embeddings.

\begin{citemize}
    \item \textbf{BERT} \cite{BERT}: We employ three pretrained models of dimension $768$:\footnote{From
    \scriptsize\url{https://github.com/google-research/bert}.}
    an English one for the English treebanks ({\footnotesize\ttfamily Base Uncased}), a~Chinese
    one for Chinese and Japanese treebanks ({\footnotesize\ttfamily Base Chinese}) and
    a multilingual one ({\footnotesize\ttfamily Base Multilingual Uncased}) for all other languages.
    We produce embedding of a UD word as an average
    of BERT subword embeddings this UD word was decomposed into, and we
    average the last four layers of the BERT model.
    \item \textbf{Flair} \cite{Akbik}: Pretrained contextual word
    embeddings of dimension $4096$ for available languages.\footnote{Models
    available in Jan 2018, for languages {\scriptsize\ttfamily bg}, {\scriptsize\ttfamily cs},
    {\scriptsize\ttfamily de}, {\scriptsize\ttfamily en}, {\scriptsize\ttfamily fr}, {\scriptsize\ttfamily nl},
    {\scriptsize\ttfamily pl}, {\scriptsize\ttfamily pt}, {\scriptsize\ttfamily sl}, {\scriptsize\ttfamily sv}.}
    \item \textbf{ELMo} \cite{Peters2018}: Pretrained contextual
    word embeddings of dimension $512$, available only for English.
\end{citemize}

We evaluate the metrics defined in \citet{CoNLL2018} using the official
evaluation
script.\footnote{\scriptsize\url{http://universaldependencies.org/conll18/conll18_ud_eval.py}}
When reporting results for multiple treebanks, we compute macro-average of
their scores (following the CoNLL 2018 Shared Task).

\section{Results}
\label{section:results}

\begin{table*}[t]
  \begin{center}
    \setlength{\tabcolsep}{5pt}
    \begin{tabular}{l|c|c|c||r|r|r|r|r|r|r|r}
\multicolumn{1}{c|}{WE} & \multicolumn{1}{c|}{CLE} & \multicolumn{1}{c|}{Bert} & \multicolumn{1}{c||}{Flair} & \multicolumn{1}{c|}{UPOS} & \multicolumn{1}{c|}{XPOS} & \multicolumn{1}{c|}{UFeats} & \multicolumn{1}{c|}{Lemmas} & \multicolumn{1}{c|}{UAS} & \multicolumn{1}{c|}{LAS} & \multicolumn{1}{c|}{MLAS} & \multicolumn{1}{c}{BLEX}\\\hline\hline
 &  &  &  & 92.77 & 89.59 & 88.88 & 91.52 & 82.59 & 77.89 & 61.52 & 65.89\\\hline
WE &  &  &  & 96.63 & 94.48 & 94.01 & 94.82 & 88.55 & 85.25 & 73.38 & 75.74\\\hline
 & CLE &  &  & 96.80 & 95.11 & 94.64 & 97.31 & 88.88 & 85.51 & 74.37 & 78.87\\\hline
WE & CLE &  &  & 97.32 & 95.88 & 95.44 & 97.62 & 89.55 & 86.46 & 76.42 & 80.36\\\hline
\hline
 &  & Base &  & 97.49 & 95.68 & 95.17 & 95.45 & 91.48 & 88.69 & 78.61 & 80.14\\\hline
WE &  & Base &  & 97.65 & 96.11 & 95.58 & 95.86 & 91.59 & 88.84 & 79.30 & 80.79\\\hline
 & CLE & Base &  & 97.79 & 96.45 & 95.94 & 97.75 & 91.74 & 88.98 & 79.97 & 83.43\\\hline
WE & CLE & Base &  & 97.89 & 96.58 & 96.09 & 97.78 & 91.80 & 89.09 & 80.30 & 83.59\\\hline
\hline
 &  &  & Flair & 97.69 & 96.22 & 95.69 & 96.49 & 90.43 & 87.57 & 77.91 & 80.06\\\hline
WE &  &  & Flair & 97.77 & 96.37 & 95.87 & 96.62 & 90.53 & 87.69 & 78.37 & 80.37\\\hline
 & CLE &  & Flair & 97.72 & 96.40 & 95.94 & 97.77 & 90.58 & 87.74 & 78.47 & 81.94\\\hline
WE & CLE &  & Flair & 97.76 & 96.50 & 96.06 & 97.85 & 90.66 & 87.83 & 78.73 & 82.16\\\hline
\hline
WE & CLE & Base & Flair & \bf 98.00 & \bf 96.80 & \bf 96.30 & \bf 97.87 & \bf 91.92 & \bf 89.32 & \bf 80.78 & \bf 83.96\\\hline
\end{tabular}

  \end{center}
  \caption{Flair compared to word embeddings (WE), character-level word
  embeddings (CLE) and BERT Base.}
  \label{table:flair}
\end{table*}

\begin{table*}[t]
  \begin{center}
    \setlength{\tabcolsep}{4pt}
    \begin{tabular}{l|c|c|c|c||r|r|r|r|r|r|r|r}
\multicolumn{1}{c|}{WE} & \multicolumn{1}{c|}{CLE} & \multicolumn{1}{c|}{Bert} & \multicolumn{1}{c|}{Flair} & \multicolumn{1}{c||}{Elmo} & \multicolumn{1}{c|}{UPOS} & \multicolumn{1}{c|}{XPOS} & \multicolumn{1}{c|}{UFeats} & \multicolumn{1}{c|}{Lemmas} & \multicolumn{1}{c|}{UAS} & \multicolumn{1}{c|}{LAS} & \multicolumn{1}{c|}{MLAS} & \multicolumn{1}{c}{BLEX}\\\hline\hline
 &  &  &  &  & 92.31 & 91.18 & 92.11 & 93.67 & 82.16 & 77.27 & 63.00 & 66.20\\\hline
WE &  &  &  &  & 95.69 & 95.30 & 96.15 & 96.27 & 86.98 & 83.59 & 73.29 & 75.40\\\hline
 & CLE &  &  &  & 95.50 & 95.04 & 95.65 & 97.06 & 86.86 & 83.10 & 72.60 & 75.53\\\hline
WE & CLE &  &  &  & 96.33 & 95.86 & 96.44 & 97.32 & 87.83 & 84.52 & 75.08 & 77.65\\\hline
\hline
 &  & Base &  &  & 96.88 & 96.46 & 96.94 & 96.18 & 90.98 & 87.98 & 79.66 & 79.94\\\hline
WE &  & Base &  &  & 97.04 & 96.66 & 97.07 & 96.38 & 91.19 & 88.20 & 80.08 & 80.41\\\hline
 & CLE & Base &  &  & 97.21 & 96.82 & 97.08 & 97.61 & 91.23 & 88.32 & 80.42 & 82.38\\\hline
WE & CLE & Base &  &  & 97.38 & 96.97 & 97.22 & 97.70 & 91.09 & 88.22 & 80.48 & 82.38\\\hline
\hline
 &  &  & Flair &  & 96.88 & 96.45 & 96.99 & 97.01 & 89.50 & 86.42 & 78.03 & 79.36\\\hline
WE &  &  & Flair &  & 97.06 & 96.56 & 97.03 & 97.12 & 89.68 & 86.67 & 78.55 & 79.85\\\hline
 & CLE &  & Flair &  & 97.00 & 96.52 & 97.04 & 97.57 & 89.75 & 86.72 & 78.56 & 80.56\\\hline
WE & CLE &  & Flair &  & 97.02 & 96.55 & 97.12 & 97.63 & 89.67 & 86.64 & 78.41 & 80.48\\\hline
\hline
 &  &  &  & Elmo & 97.23 & 96.83 & 97.25 & 97.13 & 90.15 & 87.26 & 79.47 & 80.49\\\hline
WE &  &  &  & Elmo & 97.24 & 96.84 & 97.28 & 97.12 & 90.25 & 87.34 & 79.49 & 80.57\\\hline
 & CLE &  &  & Elmo & 97.21 & 96.81 & 97.23 & 97.62 & 90.22 & 87.30 & 79.51 & 81.32\\\hline
WE & CLE &  &  & Elmo & 97.21 & 96.82 & 97.27 & 97.63 & 90.33 & 87.42 & 79.66 & 81.50\\\hline
\hline
WE & CLE & Base & Flair &  & \bf 97.45 & \bf 97.08 & 97.36 & \bf 97.76 & \bf 91.25 & \bf 88.45 & \bf 80.94 & \bf 82.79\\\hline
WE & CLE & Base &  & Elmo & 97.42 & 97.05 & 97.41 & 97.68 & 91.09 & 88.26 & 80.81 & 82.48\\\hline
WE & CLE & Base & Flair & Elmo & 97.44 & \bf 97.08 & \bf 97.43 & 97.67 & 91.08 & 88.28 & 80.76 & 82.47\\\hline
\end{tabular}

  \end{center}
  \caption{ELMo, Flair and BERT contextualized word embeddings for four
  macro-averaged English UD 2.3 treebanks. All experiments were performed
  three times and averaged.}
  \label{table:elmo}
\end{table*}

\begin{table*}[t]
  \begin{center}
    \small
    \catcode`@ = 13\def@{\bfseries}
    \catcode`! = 13\def!{\itshape}
  \setlength{\tabcolsep}{3pt}\begin{tabular}{l||r|r|r|r|r|r|r|r}
    \multicolumn{1}{c|}{System} & \multicolumn{1}{c|}{UPOS} & \multicolumn{1}{c|}{XPOS} & \multicolumn{1}{c|}{UFeats} & \multicolumn{1}{c|}{Lemmas} & \multicolumn{1}{c|}{UAS} & \multicolumn{1}{c|}{LAS} & \multicolumn{1}{c|}{MLAS} & \multicolumn{1}{c}{BLEX}\\\hline\hline

    UDPipe 2.0 WE+CLE                              &  95.84 &  94.96 &  94.24 &  95.89 &  85.53 &  82.11 &  72.12 &  75.74 \\\hline
    UDPipe 2.0 WE+CLE+BERT                         &  96.23 &  95.43 &  94.74 &  96.03 &  87.33 &  84.20 &  75.15 &  78.30 \\\hline
    UDPipe 2.0 WE+CLE+BERT 3-model ensemble        & @96.32 & @95.55 & @94.90 & @96.16 & @87.64 & @84.60 & @75.76 & @78.88 \\\hline
    \hline
    !Original UDPipe 2.0 ST entry \cite{UDPipe2.0} & !95.73 & !94.79 & !94.11 & !95.12 & !85.28 & !81.83 & !71.71 & !74.67 \\\hline
    !HIT-SCIR Harbin \cite{udst18:harbin} 3-model ensemble & !96.23 & !95.16 & !91.20 & !93.42 & !87.61 & !84.37 & !70.12 & !75.05 \\\hline
    !HIT-SCIR Harbin \cite{udst18:harbin} w/o ensembling   &        &        &        &        &        & !83.75 &        &        \\\hline
    !Stanford \cite{udst18:stanford}               & !95.93 & !94.95 & !94.14 & !95.25 & !86.56 & !83.03 & !72.67 & !75.46 \\\hline
    !TurkuNLP \cite{udst18:turku}                  & !95.41 & !94.47 & !93.82 & !96.08 & !85.32 & !81.85 & !71.27 & !75.83 \\\hline
  \end{tabular}
  \end{center}
  \caption{CoNLL 2018 UD Shared Task results on treebanks with development sets
  (so called \emph{big treebanks} in the shared task).}
  \label{table:udst}
\end{table*}


Table~\ref{table:bert} displays results for 72 UD 2.3 treebanks with train and
development sets and non-empty Wikipedia (raw corpus for the WE), considering
WE, CLE and Base BERT embeddings. Both WE and CLE bring substantial
performance boost, with CLE providing larger improvements, especially for
lemmatization and morphological features. Combining WE and CLE shows that
the improvements are complementary and using both embeddings yields
further increase.

Employing only the BERT embeddings results in significant improvements,
compared to both WE and CLE individually, with highest increase for syntactic parsing,
less for morphology and worse performance for lemmatization than CLE.
Considering BERT versus WE+CLE, BERT offers higher parsing
performance, comparable UPOS accuracy, worse morphological features and
substantially lower lemmatization performance. We therefore conclude that the
representation computed by BERT captures higher-level syntactic and possibly
even semantic meaning, while providing less information about morphology
and orthographical composition required for lemmatization.

Combining BERT and CLE results in an increased performance, especially for
morphological features and lemmatization. The addition of WE
provides minor improvements in all metrics, suggesting that the BERT embeddings
encompass substantial amount of information which WE adds to CLE.
In total, adding BERT embeddings to a baseline with WE and CLE
provides a $16.9\%$ relative error reduction for UPOS tags, $12\%$ for
morphological features, $4.3\%$ for lemmatization, and $14.5\%$ for labeled
dependency parsing.

The influence of multilingual and language-specific BERT models is analyzed in
Table~\ref{table:bert_langs}. Surprisingly, averaged results of the four
English treebanks show very little decrease when using the multilingual BERT
model compared to English-specific one, most likely owing to the fact that
English is the largest language used to train the multilingual model.
Contrary to
English, the Chinese BERT model shows substantial improvements compared to
a multilingual model when utilized on the Chinese-GSD treebank, and
minor improvements on the Japanese-GSD treebank.

Note that according to the above comparison, the substantial improvements
offered by BERT embeddings can be achieved using a \emph{single multilingual
model}, opening possibilities for interesting language-agnostic approaches.

\subsection{Flair}

Table~\ref{table:flair} shows the experiments in which WE, CLE, Flair and BERT
embeddings are added to the baseline, averaging results for 23 UD 2.3 treebanks
for which the Flair embeddings were available. 

Comparing Flair and BERT embeddings, the former demonstrates higher performance
in POS tagging, morphological features, and lemmatization, while
achieving worse results in dependency parsing, suggesting that Flair embeddings
capture more morphological and orthographical information. 
A comparison of Flair+WE+CLE with BERT+WE+CLE shows that the introduction of
WE+CLE embeddings to BERT encompasses nearly all information of Flair
embeddings, as demonstrated by BERT+WE+CLE achieving better performance in all
tasks but lemmatization, where it is only slightly behind Flair+WE+CLE.

The combination of all embeddings produces best results in all metrics.
In total, addition of BERT and Flair
embeddings to a baseline with WE and CLE provides a $25.4\%$ relative error
reduction for UPOS tags, $18.8\%$ for morphological features, $10\%$ for
lemmatization and $21\%$ for labeled dependency parsing.

\subsection{ELMo}

Given that pretrained ELMo embeddings are available for English only,
we present results for ELMo, Flair, and BERT contextualized embeddings on four
macro-averaged English UD 2.3 treebanks in Table~\ref{table:elmo}.

Flair and BERT results are consistent with the previous experiments.
Employing solely ELMo embeddings
achieves best POS tagging and lemmatization compared to using only BERT or Flair, with dependency
parsing performance higher than Flair, but lower than BERT. Therefore,
ELMo embeddings seem to encompass the most morphological and ortographical features
compared to BERT and Flair, more syntactical features than Flair, but less
than BERT.

When comparing ELMo with Flair+WE+CLE, the former surpass the latter in all
metrics but lemmatization (and lemmatization performance is equated when
employing ELMo+WE+CLE). Furthermore, morphological feature generation performance
of ELMo is better than BERT+WE+CLE. These results indicate that ELMo capture
a lot of information present in WE+CLE, which is further promoted
by the fact that ELMo+WE+CLE shows very little improvements compared to ELMo
only (with the exception of lemmatization profiting from CLE).

Overall, the best-performing model on English treebanks is
BERT+Flair+WE+CLE, with the exception of morphological features, where
ELMo helps marginally. The relative error reduction compared
to WE+CLE range from $30.5\%$ for UPOS tagging, $26\%$ for morphological features,
$16.5\%$ for lemmatization and $25.4\%$ for labeled dependency parsing.

\subsection{CoNLL 2018 Shared Task Results}

Given that the inputs in the CoNLL 2018 Shared Task are raw texts, we reuse
tokenization and segmentation employed by original UDPipe 2.0. Also, we
pretrain WE not only on Wikipedia, but on all plaintexts provided by the
shared tasks organizers. The resulting F1 scores of UDPipe 2.0 WE+CLE and WE+CLE+BERT
on treebanks with development sets (so called \emph{big treebanks} in the shared task)
are presented in Table~\ref{table:udst}.

The inclusion of BERT embeddings results in state-of-the-art single-model performance in UPOS, XPOS,
UFeats, MLAS, and BLEX metrics, and state-of-the-art ensemble performance in
all metrics. 

\subsection{BERT and Flair Improvement Levels}

\begin{figure*}[p]
  \begin{center}
    \setlength{\tabcolsep}{0pt}\begin{tabular}{cc}
      \includegraphics[width=.48\hsize]{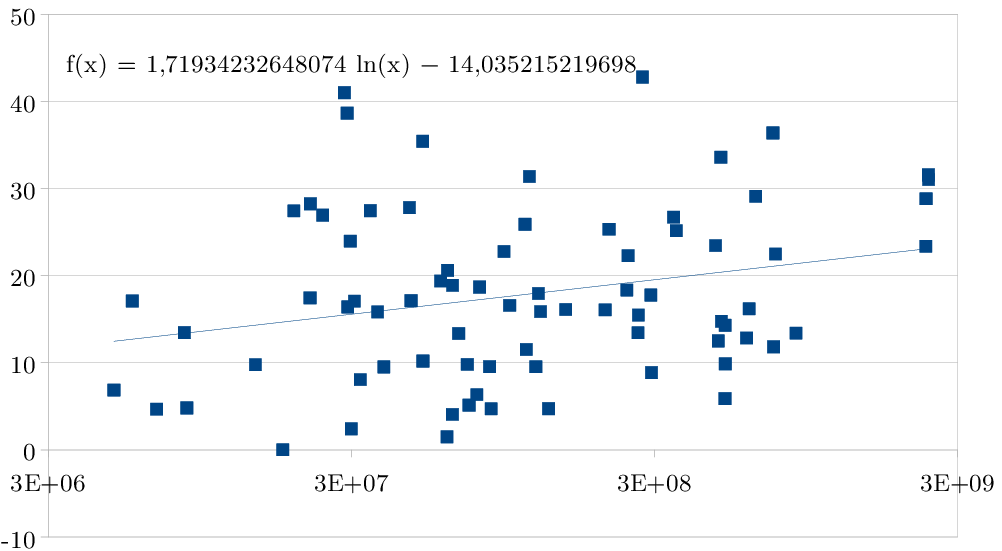} & \includegraphics[width=.48\hsize]{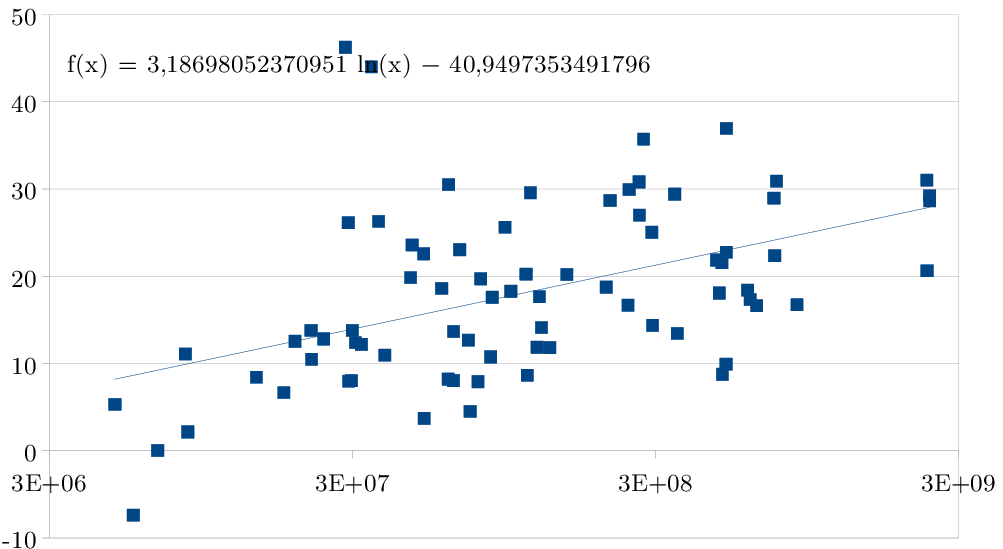} \\
      (a) UPOS & (c) UAS \\[10pt]
      \includegraphics[width=.48\hsize]{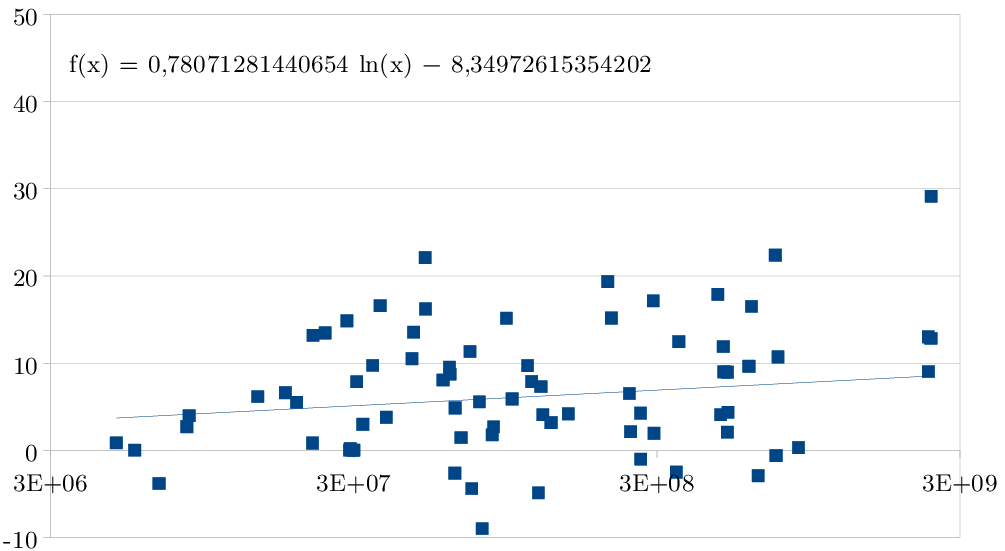} & \includegraphics[width=.48\hsize]{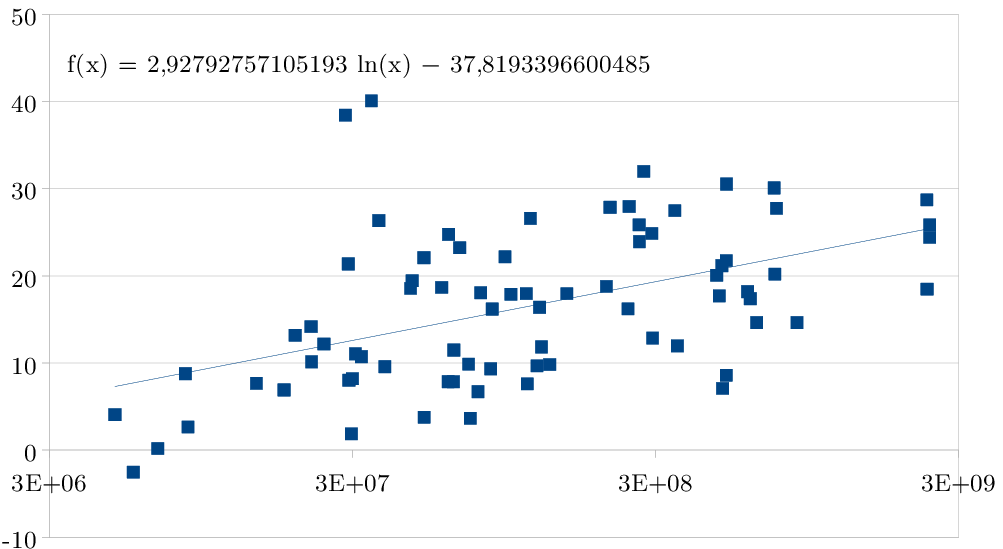} \\
      (b) Lemmas & (d) LAS \\
    \end{tabular}
  \end{center}
  \caption{Relative error improvements on UD 2.3 treebanks which have
  a training set and their language is included in BERT model. The baseline
  model uses WE and CLE, and the improved model also uses BERT Multilingual
  contextualized embeddings. The value on the $x$-axis is the logarithmic size of the
  corresponding Wikipedia, which corresponds to training data size of the BERT
  Multilingual model.}
  \label{figure:bert_wiki}

  \vspace{2.5\baselineskip}

  \begin{center}
    \setlength{\tabcolsep}{0pt}\begin{tabular}{cc}
      \includegraphics[width=.48\hsize]{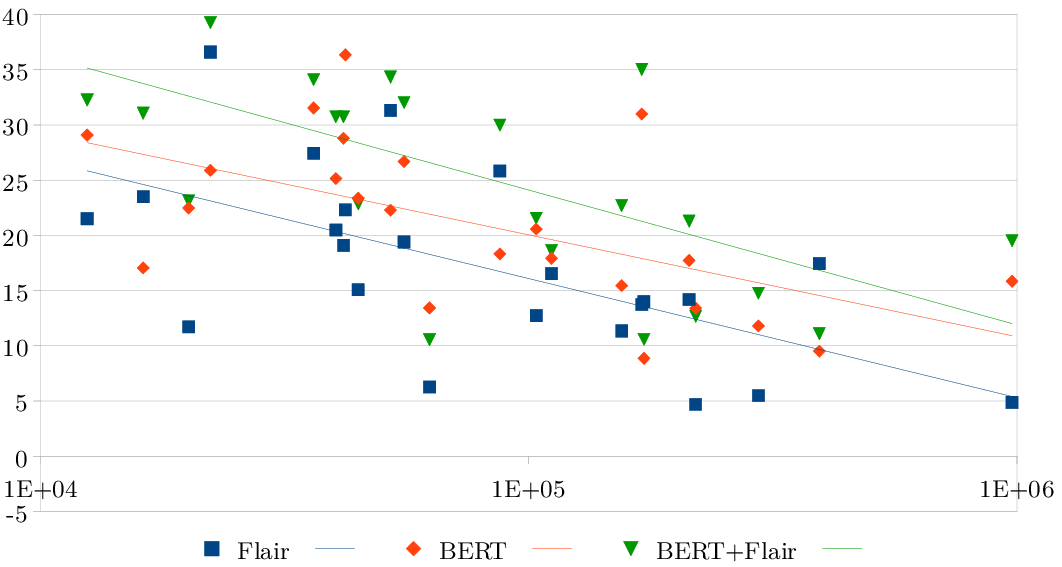} & \includegraphics[width=.48\hsize]{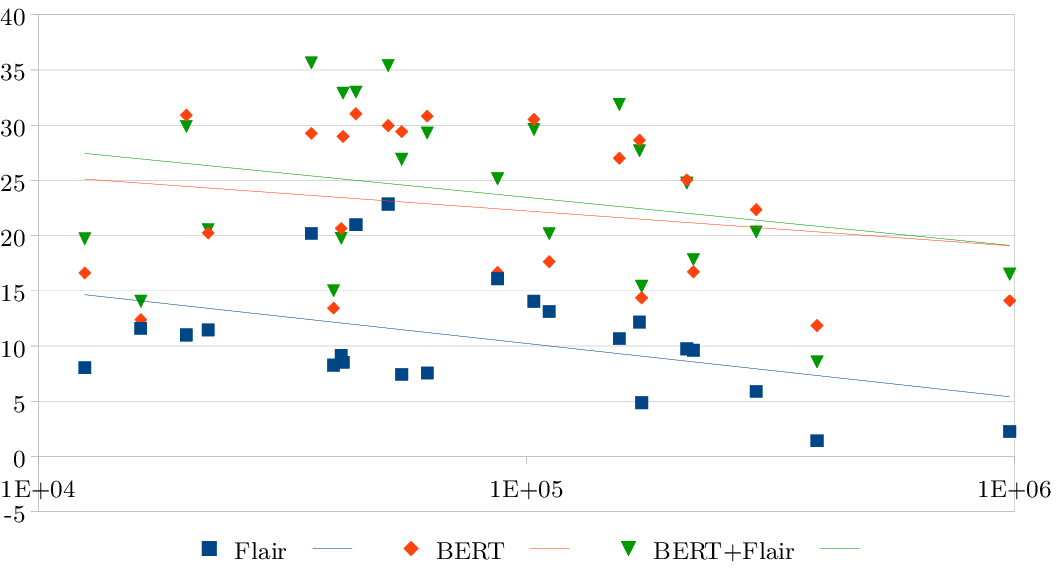} \\
      (a) UPOS & (c) UAS \\[10pt]
      \includegraphics[width=.48\hsize]{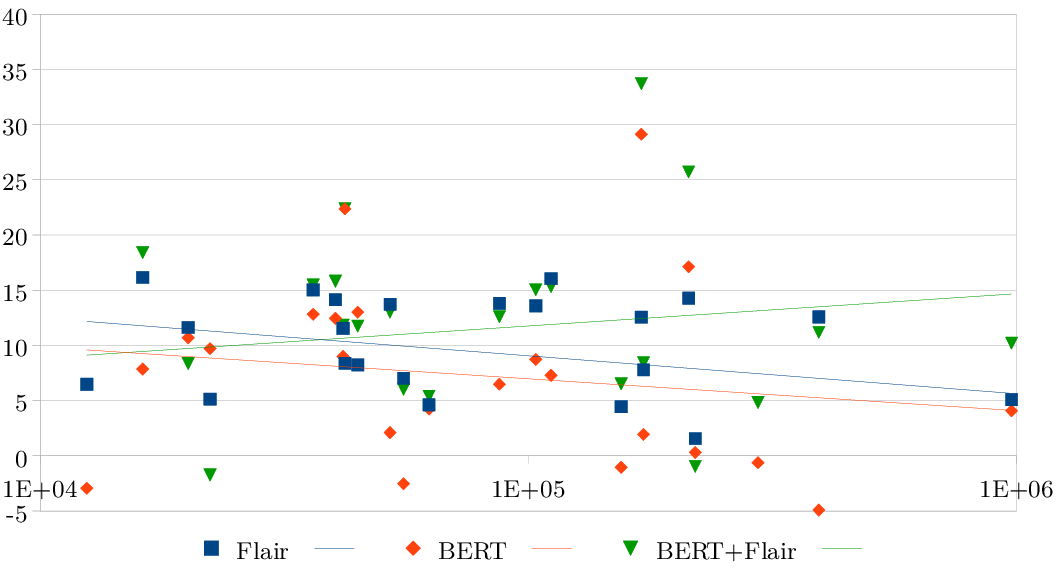} & \includegraphics[width=.48\hsize]{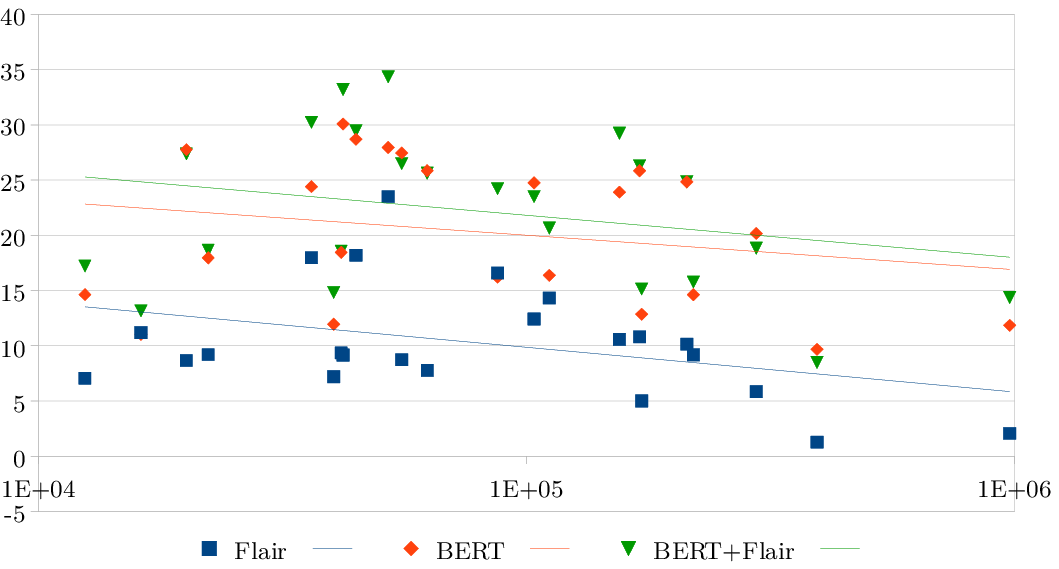} \\
      (b) Lemmas & (d) LAS \\
    \end{tabular}
  \end{center}
  \caption{Relative error improvements of the baseline with WE+CLE and
  a model additionally including Flair and/or BERT Multilingual contextual
  embeddings. The value on the $x$-axis is the logarithmic UD train data size.}
  \label{figure:flair_bert_size}
\end{figure*}

To investigate which languages benefit most from BERT embeddings,
Figure~\ref{figure:bert_wiki} presents relative error
reductions in UPOS tagging, lemmatization, and unlabeled and labeled
dependency parsing, as a function of logarithmic size of the respective Wikipedia
(which corresponds to the size of BERT Multilingual model training data).
The results indicate that consistently with intuition, larger amount of
data used to pretrain the BERT model leads to higher performance.

To compare BERT and Flair embeddings, Figure~\ref{figure:flair_bert_size}
displays relative error improvements of Flair+WE+CLE, BERT+WE+CLE and
BERT+Flair+WE+CLE models compared to WE+CLE, this time as a function of
logarithmic training data size. Generally the relative error reduction decrease
with the increasing amount of training data. Furthermore, the difference
between Flair and BERT is clearly visible, with BERT excelling in dependency
parsing and Flair in lemmatization.

\subsection{UD 2.3 Detailed Performance}

Table~\ref{table:ud} shows a detailed evaluation of all $89$ freely available
UD 2.3 treebanks with a train set, comparing the WE+CLE baseline to the best
performing WE+CLE+BERT+Flair (where Flair available) model.

The evaluation includes also $13$ treebanks whose languages are not part of BERT
Multilingual model. For these treebanks, the effect of using BERT embeddings
is mixed, as can be observed in the Table~\ref{table:ud} indicating which
UD languages were not part of BERT training. UPOS tagging, unlabeled and
labeled dependency parsing profits from BERT embedding utilization, with
averaged relative error reduction of $3.8\%$, $2\%$, and $0.8\%$, respectively.
On the other hand, lemmatization performance deteriorates, with $-2.2\%$
averaged relative error reduction.

Averaged across all
treebanks, relative error improvement of BERT+Flair embeddings inclusion is
15\% for UPOS tagging, 2.4\% for lemmatization and 11.5\% for labeled
dependency parsing.

\begin{table*}[p]
  \begin{center}
    \tiny\def\arraystretch{0.97}
    \setlength{\tabcolsep}{3pt}
    \begin{tabular}{l|c||r|r|r|r|r|r|r|r||r|r|r|r|r|r|r|r}
\multicolumn{1}{c|}{\multirow{2}{*}{Language}} & \multicolumn{1}{c||}{BERT} & \multicolumn{8}{c||}{UDPipe 2.0 with WE+CLE} & \multicolumn{8}{c}{UDPipe 2.0 with WE+CLE+BERT+Flair where available}\\\cline{3-18}
 &\multicolumn{1}{c||}{\kern.1emTrain} & \multicolumn{1}{c|}{UPOS} & \multicolumn{1}{c|}{XPOS} & \multicolumn{1}{c|}{\kern-.1emUFeats\kern-.2em} & \multicolumn{1}{c|}{\kern-.2emLemmas\kern-.2em} & \multicolumn{1}{c|}{UAS} & \multicolumn{1}{c|}{LAS} & \multicolumn{1}{c|}{MLAS} & \multicolumn{1}{c||}{BLEX} & \multicolumn{1}{c|}{UPOS} & \multicolumn{1}{c|}{XPOS} & \multicolumn{1}{c|}{\kern-.1emUFeats\kern-.2em} & \multicolumn{1}{c|}{\kern-.2emLemmas\kern-.2em} & \multicolumn{1}{c|}{UAS} & \multicolumn{1}{c|}{LAS} & \multicolumn{1}{c|}{MLAS} & \multicolumn{1}{c}{BLEX}\\\hline\hline
Afrikaans-AfriBooms &  & 98.25 & 94.48 & 97.66 & 97.46 & 89.38 & 86.58 & 77.66 & 77.82 & \bf 98.73 & \bf 95.82 & \bf 98.49 & \bf 97.60 & \bf 90.71 & \bf 88.35 & \bf 81.14 & \bf 80.17\\\hline
Ancient Greek-PROIEL & \ding{55} & \bf 97.86 & \bf 98.08 & \bf 92.44 & \bf 93.51 & \bf 85.93 & \bf 82.11 & \bf 67.16 & \bf 71.22 & 97.75 & 97.99 & 92.29 & 93.26 & 85.87 & 82.08 & 66.89 & 70.68\\\hline
Ancient Greek-Perseus & \ding{55} & \bf 93.27 & \bf 86.22 & \bf 91.39 & \bf 85.02 & \bf 78.85 & \bf 73.54 & \bf 53.87 & \bf 53.19 & 92.95 & 85.46 & 90.94 & 84.59 & 78.55 & 72.96 & 52.92 & 52.62\\\hline
Arabic-PADT &  & 96.83 & 93.97 & 94.11 & 95.28 & 87.54 & 82.94 & 73.92 & 75.87 & \bf 96.98 & \bf 94.57 & \bf 94.72 & \bf 95.43 & \bf 89.01 & \bf 84.62 & \bf 76.28 & \bf 77.81\\\hline
Armenian-ArmTDP &  & 93.49 & --- & 82.85 & 92.86 & 78.62 & 71.27 & 48.11 & 60.11 & \bf 95.30 & --- & \bf 86.89 & \bf 93.61 & \bf 82.86 & \bf 76.60 & \bf 56.15 & \bf 65.53\\\hline
Basque-BDT &  & 96.11 & --- & 92.48 & 96.29 & 86.11 & 82.86 & 72.33 & 78.54 & \bf 96.48 & --- & \bf 93.32 & \bf 96.43 & \bf 87.63 & \bf 84.50 & \bf 74.91 & \bf 80.10\\\hline
Belarusian-HSE &  & 93.63 & 89.80 & 73.30 & 87.34 & 78.58 & 72.72 & 46.20 & 58.28 & \bf 96.24 & \bf 93.27 & \bf 79.67 & \bf 89.22 & \bf 88.49 & \bf 83.21 & \bf 58.44 & \bf 69.11\\\hline
Bulgarian-BTB &  & 98.98 & 97.00 & 97.82 & 97.94 & 93.38 & 90.35 & 83.63 & 84.42 & \bf 99.20 & \bf 97.57 & \bf 98.22 & \bf 98.25 & \bf 95.34 & \bf 92.62 & \bf 87.00 & \bf 87.59\\\hline
Buryat-BDT & \ding{55} & 40.34 & --- & 32.40 & \bf 58.17 & 32.60 & \bf 18.83 & 1.26 & \bf 6.49 & \bf 45.50 & --- & \bf 33.49 & 57.42 & \bf 35.88 & 18.28 & \bf 1.48 & 5.82\\\hline
Catalan-AnCora &  & 98.88 & 98.88 & 98.37 & 99.07 & 93.22 & 91.06 & 84.48 & 86.18 & \bf 99.06 & \bf 99.06 & \bf 98.60 & \bf 99.25 & \bf 94.49 & \bf 92.74 & \bf 87.36 & \bf 88.90\\\hline
Chinese-GSD &  & 94.88 & 94.72 & 99.22 & \bf 99.99 & 84.64 & 80.50 & 71.04 & 76.78 & \bf 97.07 & \bf 96.89 & \bf 99.58 & 99.98 & \bf 90.13 & \bf 86.74 & \bf 79.67 & \bf 83.85\\\hline
Coptic-Scriptorium & \ding{55} & \bf 94.72 & \bf 93.52 & 96.27 & 95.53 & \bf 85.69 & \bf 81.08 & 64.65 & 68.65 & 94.55 & 93.15 & \bf 96.44 & \bf 95.73 & 85.10 & 80.52 & \bf 65.16 & \bf 68.81\\\hline
Croatian-SET &  & 98.13 & --- & 92.25 & 97.27 & 91.10 & 86.78 & 73.61 & 81.19 & \bf 98.45 & --- & \bf 93.27 & \bf 97.64 & \bf 93.20 & \bf 89.35 & \bf 77.08 & \bf 84.44\\\hline
Czech-CAC &  & 99.37 & 96.66 & 96.34 & 98.57 & 92.99 & 90.71 & 84.30 & 87.18 & \bf 99.44 & \bf 96.94 & \bf 96.62 & \bf 98.73 & \bf 93.59 & \bf 91.50 & \bf 85.84 & \bf 88.47\\\hline
Czech-CLTT &  & 98.88 & 91.18 & 91.59 & \bf 98.25 & 86.90 & 84.03 & 71.63 & 79.20 & \bf 99.32 & \bf 92.67 & \bf 92.88 & 98.22 & \bf 89.59 & \bf 87.01 & \bf 75.53 & \bf 82.13\\\hline
Czech-FicTree &  & 98.55 & 95.04 & 95.87 & 98.63 & 92.91 & 89.75 & 81.04 & 85.49 & \bf 98.82 & \bf 96.16 & \bf 96.88 & \bf 98.84 & \bf 94.34 & \bf 91.87 & \bf 84.80 & \bf 88.16\\\hline
Czech-PDT &  & 99.18 & 97.28 & 97.23 & 99.02 & 93.33 & 91.31 & 86.15 & 88.60 & \bf 99.34 & \bf 97.71 & \bf 97.67 & \bf 99.12 & \bf 94.43 & \bf 92.56 & \bf 88.09 & \bf 90.22\\\hline
Danish-DDT &  & 97.78 & --- & 97.33 & 97.52 & 86.88 & 84.31 & 76.29 & 78.51 & \bf 98.21 & --- & \bf 97.77 & \bf 97.72 & \bf 89.32 & \bf 87.24 & \bf 80.58 & \bf 81.93\\\hline
Dutch-Alpino &  & 96.83 & 94.80 & 96.33 & 97.09 & 91.37 & 88.38 & 77.28 & 79.82 & \bf 97.55 & \bf 95.87 & \bf 97.34 & \bf 97.28 & \bf 94.12 & \bf 91.78 & \bf 83.12 & \bf 84.42\\\hline
Dutch-LassySmall &  & 96.50 & 95.08 & 96.42 & 97.41 & 90.20 & 86.39 & 77.19 & 78.83 & \bf 96.87 & \bf 95.91 & \bf 96.97 & \bf 97.55 & \bf 93.07 & \bf 89.88 & \bf 82.00 & \bf 83.26\\\hline
English-EWT &  & 96.29 & 96.10 & 97.10 & 98.25 & 89.63 & 86.97 & 79.00 & 82.36 & \bf 97.59 & \bf 97.41 & \bf 97.82 & \bf 98.84 & \bf 92.50 & \bf 90.40 & \bf 84.41 & \bf 87.03\\\hline
English-GUM &  & 96.02 & 95.90 & 96.82 & 96.85 & 87.27 & 84.12 & 73.51 & 74.68 & \bf 96.93 & \bf 96.73 & \bf 97.59 & \bf 97.22 & \bf 91.47 & \bf 88.80 & \bf 80.14 & \bf 80.62\\\hline
English-LinES &  & 96.91 & 95.62 & 96.31 & 96.45 & 84.15 & 79.71 & 71.38 & 73.22 & \bf 97.86 & \bf 96.94 & \bf 97.48 & \bf 96.87 & \bf 87.28 & \bf 83.48 & \bf 77.45 & \bf 78.36\\\hline
English-ParTUT &  & 96.10 & 95.83 & 95.51 & 97.74 & 90.29 & 87.27 & 76.44 & 80.33 & \bf 97.43 & \bf 97.25 & \bf 96.54 & \bf 98.09 & \bf 93.75 & \bf 91.12 & \bf 81.74 & \bf 85.13\\\hline
Estonian-EDT &  & 97.64 & 98.27 & 96.23 & 95.30 & 88.00 & 85.18 & 78.72 & 78.51 & \bf 97.83 & \bf 98.36 & \bf 96.42 & \bf 95.44 & \bf 89.46 & \bf 86.77 & \bf 80.62 & \bf 80.17\\\hline
Finnish-FTB &  & 96.65 & 95.39 & 96.62 & 95.49 & 90.68 & 87.89 & 80.58 & 81.18 & \bf 96.97 & \bf 95.61 & \bf 96.73 & \bf 95.57 & \bf 91.68 & \bf 89.02 & \bf 82.25 & \bf 82.69\\\hline
Finnish-TDT &  & 97.45 & 98.12 & 95.43 & 91.45 & 89.88 & 87.46 & 80.43 & 76.64 & \bf 97.57 & \bf 98.24 & \bf 95.80 & \bf 91.68 & \bf 91.66 & \bf 89.49 & \bf 82.89 & \bf 78.57\\\hline
French-GSD &  & 97.63 & --- & 97.13 & 98.35 & 90.65 & 88.06 & 79.76 & 82.39 & \bf 97.98 & --- & \bf 97.42 & \bf 98.43 & \bf 92.55 & \bf 90.31 & \bf 82.66 & \bf 85.09\\\hline
French-ParTUT &  & 96.93 & 96.47 & 94.43 & 95.70 & 92.17 & 89.63 & 75.22 & 78.07 & \bf 97.64 & \bf 97.35 & \bf 95.12 & \bf 96.06 & \bf 94.51 & \bf 92.47 & \bf 80.50 & \bf 82.19\\\hline
French-Sequoia &  & 98.79 & --- & 98.09 & 98.57 & 92.37 & 90.73 & 84.51 & 85.93 & \bf 99.32 & --- & \bf 98.62 & \bf 98.89 & \bf 94.88 & \bf 93.81 & \bf 89.10 & \bf 90.08\\\hline
French-Spoken &  & 95.91 & 97.30 & --- & \bf 96.92 & 82.90 & 77.53 & 68.24 & 69.47 & \bf 97.23 & \bf 97.48 & --- & 96.75 & \bf 86.27 & \bf 81.40 & \bf 73.26 & \bf 73.36\\\hline
Galician-CTG &  & 97.84 & 97.47 & \bf 99.83 & 98.58 & 86.44 & 83.82 & 72.46 & 77.21 & \bf 98.06 & \bf 97.70 & \bf 99.83 & \bf 98.81 & \bf 86.94 & \bf 84.43 & \bf 73.72 & \bf 78.33\\\hline
Galician-TreeGal &  & 95.82 & 92.46 & 93.96 & 97.06 & 82.72 & 77.69 & 63.73 & 68.89 & \bf 97.30 & \bf 95.01 & \bf 96.03 & \bf 97.71 & \bf 86.62 & \bf 82.62 & \bf 72.29 & \bf 76.24\\\hline
German-GSD &  & 94.48 & 97.31 & 90.68 & \bf 96.80 & 85.53 & 81.07 & 58.82 & 72.13 & \bf 95.18 & \bf 97.95 & \bf 91.72 & 96.77 & \bf 88.11 & \bf 84.06 & \bf 63.33 & \bf 75.44\\\hline
Gothic-PROIEL & \ding{55} & 96.66 & \bf 97.23 & \bf 90.77 & \bf 94.72 & 85.27 & 79.60 & 66.71 & \bf 72.86 & \bf 96.72 & 97.22 & 90.58 & 94.47 & \bf 85.53 & \bf 79.69 & \bf 66.86 & 72.52\\\hline
Greek-GDT &  & 97.98 & 97.99 & 94.96 & 95.82 & 92.10 & 89.79 & 78.60 & 79.72 & \bf 98.25 & \bf 98.25 & \bf 95.76 & \bf 95.88 & \bf 93.92 & \bf 92.16 & \bf 82.29 & \bf 82.14\\\hline
Hebrew-HTB &  & 97.02 & 97.03 & 95.87 & 97.12 & 89.70 & 86.86 & 75.52 & 78.14 & \bf 97.50 & \bf 97.50 & \bf 96.18 & \bf 97.24 & \bf 91.78 & \bf 89.22 & \bf 78.85 & \bf 80.80\\\hline
Hindi-HDTB &  & 97.52 & 97.04 & 94.15 & \bf 98.67 & 94.85 & 91.83 & 78.49 & 86.83 & \bf 97.58 & \bf 97.19 & \bf 94.24 & \bf 98.67 & \bf 95.56 & \bf 92.50 & \bf 79.32 & \bf 87.66\\\hline
Hungarian-Szeged &  & 95.76 & --- & 91.75 & 95.05 & 84.04 & 79.73 & 67.63 & 73.63 & \bf 97.09 & --- & \bf 93.41 & \bf 95.44 & \bf 88.76 & \bf 85.12 & \bf 74.08 & \bf 79.21\\\hline
Indonesian-GSD &  & 93.69 & 94.19 & 95.58 & 99.64 & 85.31 & 78.99 & 67.74 & 76.38 & \bf 94.09 & \bf 94.93 & \bf 96.03 & \bf 99.66 & \bf 86.47 & \bf 80.40 & \bf 70.01 & \bf 78.19\\\hline
Irish-IDT &  & 92.72 & 91.44 & 82.43 & 90.48 & 80.39 & 72.34 & 46.49 & 55.32 & \bf 93.22 & \bf 92.00 & \bf 83.78 & \bf 90.56 & \bf 81.43 & \bf 73.47 & \bf 49.05 & \bf 56.50\\\hline
Italian-ISDT &  & 98.39 & 98.30 & 98.11 & 98.66 & 93.49 & 91.54 & 84.28 & 85.49 & \bf 98.62 & \bf 98.54 & \bf 98.26 & \bf 98.78 & \bf 94.97 & \bf 93.38 & \bf 87.14 & \bf 88.10\\\hline
Italian-ParTUT &  & 98.38 & 98.35 & 97.77 & 98.16 & 92.64 & 90.47 & 81.87 & 82.99 & \bf 98.54 & \bf 98.52 & \bf 98.05 & \bf 98.24 & \bf 95.36 & \bf 93.38 & \bf 86.57 & \bf 87.30\\\hline
Italian-PoSTWITA &  & 96.61 & 96.43 & 96.90 & 97.00 & 86.03 & 81.78 & 72.88 & 74.33 & \bf 97.11 & \bf 96.98 & \bf 97.12 & \bf 97.27 & \bf 87.25 & \bf 83.07 & \bf 74.70 & \bf 76.27\\\hline
Japanese-GSD &  & 98.13 & 97.81 & \bf 99.98 & 99.52 & 95.06 & 93.73 & 86.37 & 88.04 & \bf 98.24 & \bf 97.89 & \bf 99.98 & \bf 99.53 & \bf 95.55 & \bf 94.27 & \bf 87.64 & \bf 89.24\\\hline
Kazakh-KTB &  & 55.84 & 52.06 & 40.40 & 63.96 & 53.30 & 33.38 & 4.82 & 15.10 & \bf 63.08 & \bf 60.63 & \bf 43.64 & \bf 64.03 & \bf 57.02 & \bf 38.72 & \bf 7.88 & \bf 18.78\\\hline
Korean-GSD &  & 96.29 & 90.39 & 99.77 & 93.40 & 87.70 & 84.24 & 79.74 & 76.35 & \bf 96.99 & \bf 91.21 & \bf 99.83 & \bf 93.72 & \bf 89.38 & \bf 86.05 & \bf 82.19 & \bf 78.58\\\hline
Korean-Kaist &  & 95.59 & 87.00 & --- & \bf 94.30 & 88.42 & 86.48 & 80.72 & 79.22 & \bf 95.77 & \bf 87.46 & --- & 94.15 & \bf 89.35 & \bf 87.54 & \bf 82.12 & \bf 80.18\\\hline
Kurmanji-MG & \ding{55} & 53.38 & 51.42 & 41.53 & \bf 69.58 & \bf 45.22 & \bf 34.32 & 2.74 & \bf 19.39 & \bf 58.78 & \bf 56.11 & \bf 42.03 & 68.21 & 43.74 & 32.99 & \bf 3.10 & 17.98\\\hline
Latin-ITTB &  & 98.34 & 96.37 & 96.97 & 98.99 & 91.06 & 88.80 & 82.35 & 85.71 & \bf 98.42 & \bf 96.45 & \bf 97.05 & \bf 99.03 & \bf 91.25 & \bf 89.10 & \bf 82.80 & \bf 86.05\\\hline
Latin-PROIEL &  & 97.01 & 97.15 & 91.53 & \bf 96.32 & \bf 83.34 & 78.66 & \bf 67.40 & \bf 73.65 & \bf 97.15 & \bf 97.21 & \bf 91.54 & 96.18 & \bf 83.34 & \bf 78.70 & 67.29 & 73.52\\\hline
Latin-Perseus &  & 88.40 & 74.58 & 79.10 & 81.45 & 71.20 & 61.28 & 41.58 & 45.09 & \bf 89.96 & \bf 76.22 & \bf 80.43 & \bf 81.95 & \bf 74.39 & \bf 64.68 & \bf 44.96 & \bf 47.94\\\hline
Latvian-LVTB &  & \bf 96.11 & 88.69 & 93.01 & 95.46 & 87.20 & 83.35 & 71.92 & 76.64 & \bf 96.11 & \bf 89.06 & \bf 93.30 & \bf 95.76 & \bf 88.05 & \bf 84.50 & \bf 73.81 & \bf 78.33\\\hline
Lithuanian-HSE &  & 81.70 & 79.91 & 60.47 & \bf 76.89 & 51.98 & 42.17 & 18.17 & 28.70 & \bf 88.77 & \bf 86.04 & \bf 66.70 & \bf 76.89 & \bf 64.53 & \bf 54.53 & \bf 26.35 & \bf 34.76\\\hline
Maltese-MUDT & \ding{55} & 95.99 & 95.69 & --- & --- & 84.65 & 79.71 & 66.75 & 71.49 & \bf 96.15 & \bf 95.85 & --- & --- & \bf 85.31 & \bf 80.10 & \bf 67.21 & \bf 71.62\\\hline
Marathi-UFAL &  & 80.10 & --- & 67.23 & \bf 81.31 & \bf 70.63 & \bf 61.41 & 29.34 & \bf 45.87 & \bf 83.50 & --- & \bf 67.96 & \bf 81.31 & 68.45 & 60.44 & \bf 29.58 & 43.75\\\hline
North Sami-Giella & \ding{55} & 92.61 & 93.78 & \bf 90.00 & \bf 88.34 & 78.39 & 73.60 & 62.29 & 61.45 & \bf 92.76 & \bf 94.11 & 89.83 & 88.25 & \bf 78.47 & \bf 73.95 & \bf 62.47 & \bf 61.68\\\hline
Norwegian-Bokmaal &  & 98.31 & --- & 97.14 & 98.64 & 92.39 & 90.49 & 84.06 & 86.53 & \bf 98.59 & --- & \bf 97.54 & \bf 98.72 & \bf 93.78 & \bf 92.19 & \bf 86.72 & \bf 88.60\\\hline
Norwegian-Nynorsk &  & 93.87 & --- & 91.57 & 96.06 & 80.09 & 75.04 & 63.72 & 68.22 & \bf 95.52 & --- & \bf 93.17 & \bf 96.59 & \bf 82.64 & \bf 78.08 & \bf 67.53 & \bf 71.75\\\hline
Norwegian-NynorskLIA &  & 89.59 & --- & 86.13 & 93.93 & 68.08 & 60.07 & 44.47 & 50.98 & \bf 92.53 & --- & \bf 88.96 & \bf 94.73 & \bf 71.42 & \bf 64.12 & \bf 49.10 & \bf 55.36\\\hline
Old Church Slavonic-PROIEL & \ding{55} & 96.89 & \bf 97.16 & \bf 90.72 & \bf 93.07 & 89.64 & 84.99 & 73.66 & 77.71 & \bf 96.96 & 97.13 & 90.45 & 92.91 & \bf 89.88 & \bf 85.21 & \bf 73.77 & \bf 77.88\\\hline
Old French-SRCMF & \ding{55} & 96.09 & 96.00 & 97.82 & --- & 91.75 & \bf 86.82 & \bf 79.89 & \bf 83.81 & \bf 96.26 & \bf 96.21 & \bf 97.89 & --- & \bf 91.83 & 86.75 & 79.79 & 83.55\\\hline
Persian-Seraji &  & 97.75 & 97.70 & 97.78 & \bf 97.44 & 90.05 & 86.66 & 81.23 & 80.93 & \bf 98.17 & \bf 98.05 & \bf 98.13 & 97.21 & \bf 92.01 & \bf 89.07 & \bf 84.36 & \bf 83.40\\\hline
Polish-LFG &  & 98.80 & 94.56 & 95.49 & 97.54 & 96.58 & 94.76 & 87.04 & 90.26 & \bf 99.16 & \bf 95.91 & \bf 96.57 & \bf 97.85 & \bf 97.44 & \bf 96.03 & \bf 90.14 & \bf 92.09\\\hline
Polish-SZ &  & 98.34 & 93.25 & 93.04 & 97.16 & 93.39 & 91.24 & 81.06 & 85.99 & \bf 98.91 & \bf 95.12 & \bf 95.08 & \bf 97.53 & \bf 95.73 & \bf 94.25 & \bf 86.66 & \bf 89.89\\\hline
Portuguese-Bosque &  & 97.07 & --- & 96.40 & 98.46 & 91.36 & 89.04 & 76.67 & 83.06 & \bf 97.38 & --- & \bf 96.96 & \bf 98.59 & \bf 92.69 & \bf 90.70 & \bf 79.59 & \bf 85.44\\\hline
Portuguese-GSD &  & 98.31 & 98.30 & 99.92 & 99.30 & 93.01 & 91.63 & 85.96 & 86.94 & \bf 98.67 & \bf 98.67 & \bf 99.93 & \bf 99.48 & \bf 94.74 & \bf 93.71 & \bf 89.19 & \bf 90.28\\\hline
Romanian-Nonstandard &  & 96.68 & 92.11 & 90.88 & \bf 94.78 & 89.12 & 84.20 & 65.93 & 73.44 & \bf 96.85 & \bf 92.27 & \bf 91.04 & 94.55 & \bf 89.61 & \bf 84.78 & \bf 66.82 & \bf 73.77\\\hline
Romanian-RRT &  & 97.96 & 97.43 & 97.53 & 98.41 & 91.31 & 86.74 & 79.02 & 81.09 & \bf 98.16 & \bf 97.56 & \bf 97.75 & \bf 98.59 & \bf 92.41 & \bf 88.05 & \bf 81.04 & \bf 82.89\\\hline
Russian-GSD &  & 97.10 & 96.98 & 92.66 & 97.37 & 88.15 & 84.37 & 74.07 & 80.03 & \bf 97.78 & \bf 97.64 & \bf 94.76 & \bf 97.84 & \bf 90.74 & \bf 87.51 & \bf 79.13 & \bf 83.97\\\hline
Russian-SynTagRus &  & 99.12 & --- & 97.57 & 98.53 & 93.80 & 92.32 & 87.91 & 89.17 & \bf 99.23 & --- & \bf 97.97 & \bf 98.59 & \bf 94.92 & \bf 93.68 & \bf 89.85 & \bf 90.81\\\hline
Russian-Taiga &  & 93.18 & \bf 99.98 & 82.87 & 89.99 & 75.45 & 69.11 & 48.81 & 57.21 & \bf 95.47 & \bf 99.98 & \bf 86.87 & \bf 91.18 & \bf 80.74 & \bf 75.65 & \bf 57.16 & \bf 63.65\\\hline
Serbian-SET &  & 98.33 & --- & 94.35 & 97.36 & 92.70 & 89.27 & 79.14 & 84.18 & \bf 98.71 & --- & \bf 95.79 & \bf 97.76 & \bf 94.57 & \bf 91.65 & \bf 83.03 & \bf 87.24\\\hline
Slovak-SNK &  & 96.83 & 86.14 & 90.82 & 96.40 & 89.82 & 86.90 & 74.00 & 81.37 & \bf 97.70 & \bf 88.54 & \bf 93.07 & \bf 96.75 & \bf 94.30 & \bf 92.15 & \bf 81.43 & \bf 87.24\\\hline
Slovenian-SSJ &  & 98.61 & 95.70 & 95.92 & 98.25 & 92.96 & 91.16 & 83.85 & 86.89 & \bf 98.83 & \bf 96.53 & \bf 96.77 & \bf 98.54 & \bf 94.81 & \bf 93.49 & \bf 87.58 & \bf 90.04\\\hline
Slovenian-SST &  & 93.79 & 86.12 & 86.28 & 95.17 & 73.51 & 67.51 & 52.67 & 60.32 & \bf 95.72 & \bf 89.25 & \bf 89.43 & \bf 96.06 & \bf 77.23 & \bf 71.79 & \bf 58.69 & \bf 64.84\\\hline
Spanish-AnCora &  & 98.91 & 98.92 & 98.49 & 99.17 & 92.34 & 90.26 & 83.97 & 85.51 & \bf 99.05 & \bf 99.06 & \bf 98.70 & \bf 99.25 & \bf 93.75 & \bf 92.03 & \bf 87.03 & \bf 88.35\\\hline
Spanish-GSD &  & 96.85 & --- & 97.09 & 98.97 & 90.71 & 88.03 & 75.98 & 81.47 & \bf 97.36 & --- & \bf 97.19 & \bf 99.14 & \bf 92.32 & \bf 90.11 & \bf 79.29 & \bf 84.92\\\hline
Swedish Sign Language-SSLC & \ding{55} & 68.44 & 57.27 & --- & --- & 49.82 & 37.94 & 31.34 & 39.47 & \bf 72.34 & \bf 70.92 & --- & --- & \bf 56.03 & \bf 42.02 & \bf 34.50 & \bf 43.19\\\hline
Swedish-LinES &  & 96.78 & 94.75 & 89.43 & 97.03 & 86.07 & 81.86 & 66.48 & 77.38 & \bf 97.77 & \bf 95.97 & \bf 90.39 & \bf 97.50 & \bf 88.16 & \bf 84.55 & \bf 70.13 & \bf 80.81\\\hline
Swedish-Talbanken &  & 97.94 & 96.71 & 96.86 & 98.01 & 89.63 & 86.61 & 79.67 & 82.26 & \bf 98.60 & \bf 97.62 & \bf 97.69 & \bf 98.13 & \bf 92.42 & \bf 90.16 & \bf 84.56 & \bf 86.19\\\hline
Tamil-TTB &  & 91.05 & 83.81 & 87.28 & 93.92 & 74.11 & 66.37 & 55.31 & 59.58 & \bf 92.61 & \bf 86.53 & \bf 89.89 & \bf 93.97 & \bf 77.68 & \bf 71.14 & \bf 60.67 & \bf 64.74\\\hline
Telugu-MTG &  & 93.07 & 93.07 & \bf 99.03 & --- & 91.26 & 85.02 & 77.75 & \bf 81.76 & \bf 94.73 & \bf 94.73 & \bf 99.03 & --- & \bf 91.96 & \bf 85.30 & \bf 77.79 & 81.60\\\hline
Turkish-IMST &  & 96.01 & 95.12 & 92.55 & 96.01 & 74.19 & 67.56 & 56.96 & 61.37 & \bf 96.07 & \bf 95.37 & \bf 93.25 & \bf 96.39 & \bf 76.30 & \bf 70.11 & \bf 59.91 & \bf 64.07\\\hline
Ukrainian-IU &  & 97.59 & 92.66 & 92.66 & 97.23 & 88.29 & 85.25 & 73.81 & 79.10 & \bf 98.20 & \bf 94.63 & \bf 94.43 & \bf 97.65 & \bf 91.65 & \bf 89.36 & \bf 79.97 & \bf 84.24\\\hline
Upper Sorbian-UFAL & \ding{55} & 62.93 & --- & 41.10 & \bf 68.68 & 45.58 & 34.54 & 3.37 & 16.65 & \bf 69.69 & --- & \bf 43.46 & 66.80 & \bf 48.64 & \bf 38.85 & \bf 5.03 & \bf 17.80\\\hline
Urdu-UDTB &  & 93.66 & 91.98 & 81.92 & 97.40 & 87.50 & 81.62 & 55.02 & 73.07 & \bf 94.28 & \bf 92.37 & \bf 82.47 & \bf 97.56 & \bf 88.55 & \bf 83.03 & \bf 56.58 & \bf 75.05\\\hline
Uyghur-UDT & \ding{55} & \bf 89.87 & \bf 92.54 & \bf 88.30 & \bf 95.31 & 78.46 & 67.09 & 47.84 & 57.08 & 89.58 & 92.27 & 88.29 & 95.30 & \bf 79.10 & \bf 67.46 & \bf 48.09 & \bf 57.69\\\hline
Vietnamese-VTB &  & 89.68 & 87.41 & \bf 99.72 & 99.55 & 70.38 & 62.56 & 55.56 & 59.54 & \bf 90.87 & \bf 88.87 & 99.68 & \bf 99.79 & \bf 72.94 & \bf 65.41 & \bf 58.97 & \bf 62.64\\\hline
\hline
Total &  & 93.71 & 92.52 & 90.56 & 94.35 & 84.23 & 79.59 & 67.36 & 72.05 & \bf 94.71 & \bf 93.69 & \bf 91.50 & \bf 94.51 & \bf 86.34 & \bf 82.01 & \bf 70.66 & \bf 74.75\\\hline
\end{tabular}

  \end{center}
  \caption{Results on all UD 2.3 treebanks with a train set, comparing
  inclusion of BERT and possibly Flair embeddings to WE+CLE baseline. Gold
  tokenization and segmentation is used.}
  \label{table:ud}
\end{table*}

\section{Conclusions}

We presented a thorough evaluation of the BERT, Flair, and ELMo contextualized
embeddings in $89$ languages of the UD in POS tagging, lemmatization, and
dependency parsing. We conclude that addition of any of the contextualized
embeddings as additional inputs to a neural network results in substantial
performance increase. Our findings show that the BERT embeddings yield the
greatest improvements, reaching state-of-the-art results in CoNLL 2018 Shared
Task
and contain most complementary information
as compared to word- and character-level word embeddings,
while Flair
embeddings encompass the morphological and orthographical information.

\section*{Acknowledgements}

The work described herein has been supported by OP VVV VI LINDAT/CLARIN project
of the Ministry of Education, Youth and Sports of the Czech Republic (project
CZ.02.1.01/0.0/0.0/16\_013/0001781) and it has been supported and has been
using language resources developed by the LINDAT/CLARIN project of the the
Ministry of Education, Youth and Sports of the Czech Republic (project
LM2015071).

\bibliography{emnlp-ijcnlp-2019}
\bibliographystyle{acl_natbib}

\end{document}